\documentclass[10pt,twocolumn,letterpaper]{article}

\usepackage{cvpr}              

\usepackage{graphicx}
\usepackage{amsmath}
\usepackage{amssymb}
\usepackage{booktabs}
\usepackage{algpseudocode}
\usepackage{algorithmicx,algorithm}
\usepackage{float}
\usepackage{multirow}
\usepackage{pifont}
\usepackage{enumitem}
\usepackage{siunitx}
\usepackage{arydshln}
\usepackage[accsupp]{axessibility}
\usepackage[normalem]{ulem}
\useunder{\uline}{\ul}{}
\usepackage{xcolor}
\newsavebox\CBox
\def\textBF#1{\sbox\CBox{#1}\resizebox{\wd\CBox}{\ht\CBox}{\textbf{#1}}}
\newcommand{\bestscore}[1]{\textcolor{red}{\textBF{#1}}}
\newcommand{\secondscore}[1]{\textcolor{blue}{\underline{#1}}}

\usepackage[pagebackref,breaklinks,colorlinks]{hyperref}

\usepackage[capitalize]{cleveref}
\crefname{section}{Sec.}{Secs.}
\Crefname{section}{Section}{Sections}
\Crefname{table}{Table}{Tables}
\crefname{table}{Tab.}{Tabs.}

\def\cvprPaperID{7208} 
\def\confName{CVPR}
\def\confYear{2023}

\begin{document}

\title{Visibility Constrained Wide-band Illumination Spectrum Design for Seeing-in-the-Dark}

\author{Muyao Niu, Zhuoxiao Li, Zhihang Zhong, Yinqiang Zheng\thanks{Corresponding author} \\
The University of Tokyo \\
{\tt\small muyao.niu@gmail.com, \{lizhuoxiao@g.ecc, zhong@is.s, yqzheng@ai\}.u-tokyo.ac.jp}
}

\maketitle

\begin{abstract}
   Seeing-in-the-dark is one of the most important and challenging computer vision tasks due to its wide applications and extreme complexities of in-the-wild scenarios. Existing arts can be mainly divided into two threads: 1) RGB-dependent methods restore information using degraded RGB inputs only (\eg, low-light enhancement), 2) RGB-independent methods translate images captured under auxiliary near-infrared (NIR) illuminants into RGB domain (\eg, NIR2RGB translation). The latter is very attractive since it works in complete darkness and the illuminants are visually friendly to naked eyes, but tends to be unstable due to its intrinsic ambiguities. In this paper, we try to robustify NIR2RGB translation by designing the optimal spectrum of auxiliary illumination in the wide-band VIS-NIR range, while keeping visual friendliness. Our core idea is to quantify the visibility constraint implied by the human vision system and incorporate it into the design pipeline. By modeling the formation process of images in the VIS-NIR range, the optimal multiplexing of a wide range of LEDs is automatically designed in a fully differentiable manner, within the feasible region defined by the visibility constraint. We also collect a substantially expanded VIS-NIR hyperspectral image dataset for experiments by using a customized 50-band filter wheel. Experimental results show that the task can be significantly improved by using the optimized wide-band illumination than using NIR only. Codes Available: \url{https://github.com/MyNiuuu/VCSD}.
\end{abstract}

\section{Introduction}
Seeing-in-the-dark is critical for modern industries, because of its promising applications in nighttime photography and visual surveillance. However, it remains challenging due to complex degradation mechanisms and dynamics of in-the-wild environments. 

To achieve this task, a number of methods have been proposed, which can be roughly divided into two threads. The first thread features RGB-dependent methods~\cite{lore2017llnet,wang2019underexposed,wei2018deep,chen2018learning,wang2020practical,chen2019seeing,yue2020supervised} that aim to fully exploit the RGB input, even with severe degradations. These methods have gained great success through directly learning the mapping from low-light input to normal-light output, in the presence of complex noises and color discrepancies. However, even state-of-the-art methods along this thread may struggle with in-the-wild data captured under nearly complete darkness. 

In contrast, the second thread features RGB-independent methods~\cite{Liu_2022_CVPR, limmer2016infrared,suarez2017infrared,Wang_2020_VCIP_Semantic,nyberg2018unpaired} for non-interfering surveillance that try to recover RGB information from images of invisible ranges, without requiring any RGB input. The most attractive characteristics lie in its applicability to complete darkness and the visual friendliness of auxiliary illumination to naked eyes. NIR2RGB is one of the representative tasks of this thread, which aims to translate near-infrared images to RGB images. 

As for auxiliary illumination in the NIR range, the industry practice is to use NIR LEDs, usually centered at \SI{850}{nm} or \SI{940}{nm}. However, the captured images are almost monochromatic and lack visual color and texture, which makes NIR2RGB translation ambiguous. The fundamental reasons for the ambiguities are two folds: 1) The spectral sensitivities of commodity RGB cameras almost overlap around both \SI{850}{nm} and \SI{940}{nm}, making it hard to recover three-channel color from a single intensity observation. 2) Reflectance spectra of many materials become almost indistinguishable beyond \SI{850}{nm}, which leads to obvious structure gaps from RGB images.
As a result, existing studies that tried to directly convert such NIR images to VIS images, even with the most advanced deep learning techniques, can hardly provide satisfying results due to these fundamental restrictions. In \cite{Liu_2022_CVPR}, Liu~\etal proposed to properly multiplex different NIR LEDs, ranging from \SI{700}{nm} to \SI{1000}{nm}, to robustify the NIR2RGB task, and achieves apparently better results than using traditional \SI{850}{nm} or \SI{940}{nm} LEDs. However, structure gaps still exist due to the restriction of wavelengths in the NIR range, making the results far from satisfying.

The basic motivation of these methods arises from the invisibility of human naked eyes to NIR lights, so as to reduce visual interference and light pollution. However, up to now, none of these works have explicitly formulated the visibility of certain illumination. Liu~\etal~\cite{Liu_2022_CVPR} empirically picked up the NIR range beyond \SI{700}{nm}, and there is a clear tendency that LEDs closer to this prescribed boundary are preferred according to their results. A natural question is: Is there an exact boundary between visible and invisible? This is important since it determines how much information in the VIS range can be utilized to help RGB recovery. 

Inspired by the aforementioned methods, we propose to quantify and incorporate the human vision system into our model, which enables us to significantly robustify this task via illumination spectrum design in the wide-band spectral range from \SI{420}{nm} to \SI{890}{nm}. Similar to~\cite{Liu_2022_CVPR}, we directly optimize the spectral curve by training an image enhancement model on hyperspectral datasets. Specifically, based on the human vision system, we establish a Visibility Constrained Spectrum Design (VCSD) model to quantify the visibility of certain spectra, and to assure the prescribed visibility level will not be violated. To achieve this, a visibility threshold $\hat{\Psi}$ is introduced, which serves as the visibility upper bound during the spectrum design process. In practice, this threshold can be changed according to the desired level of visibility, without destroying the validity of our method. According to the upper bounded visibility level, the model scales down the designed LED spectrum (if necessary) to assure that the new spectrum is friendly to naked eyes. After that, we design a physic-based Imaging Process Simulation (IPS) model which synthesizes images using the corresponding LED spectrum, camera spectral sensitivity, and the reflectance spectrum of the scene. The IPS model also contains a noise model to consider the noise effect during the realistic imaging process. Since we consider the spectrum from \SI{420}{nm} to \SI{890}{nm}, we synthesize one VIS image with lights shorter than \SI{700}{nm} and one VIS-NIR image with the full spectrum. Through deep learning, we directly minimize the reconstruction loss and finally get the optimal LED spectral curve that can be physically realized by driving LEDs with appropriate voltage and current. 

We evaluate the effectiveness of our model and designed curve on hyperspectral datasets including our proposed and previous~\cite{monno2018single} datasets. Compared to existing methods, our model clearly achieves superior results, demonstrating the powerfulness of wide-band illumination spectrum design under visibility constraints.

The main highlights of this work are:
\begin{itemize}
    \item For the first time, we propose a paradigm that quantifies and incorporates the human vision system for seeing-in-the-dark, which enables us to significantly improve the task via illumination spectrum design in a wide-band coverage from \SI{420}{nm} to \SI{890}{nm}.
    \item A novel Visibility Constrained Spectrum Design (VCSD) model is proposed to formulate and assure the visibility level of certain spectra to human naked eyes during the optimization process. The visibility threshold can be changed according to the desired level of visibility, without destroying the validity of the model.
    \item We design a physic-based Imaging Process Simulation (IPS) module which synthesizes the input images based on the imaging process and the noise model.
    \item We contribute a VIS-NIR wide-band hyperspectral image dataset to supplement existing ones in terms of quality and quantity. 
\end{itemize}

\section{Related Work}

\noindent \textbf{Image Enhancement.}
Low-light image enhancement in the visible range is a critical and challenging task. Traditional image enhancement methods were mostly based on histogram manipulation~\cite{abdullah2007dynamic,coltuc2006exact,ibrahim2007brightness,lee2013contrast,stark2000adaptive} or Retinex theory~\cite{land1977retinex,jobson1997multiscale,wang2013naturalness,fu2016weighted,guo2016lime,li2018structure}. In recent years, many learning-based methods have been proposed and attracted increasingly wide interest~\cite{lore2017llnet,wang2019underexposed,wei2018deep,Zero-DCE,jiang2021enlightengan}. Wei \etal~\cite{wei2018deep} combined traditional Retinex theory with deep neural networks and provided an end-to-end framework for low-light enhancement. Supervised learning has been extensively exploited for enhancing low-light RAW images~\cite{chen2018learning,wang2020practical} and videos~\cite{chen2019seeing,jiang2019learning,yue2020supervised,Wang_2021_ICCV}. Very recently, enhancement methods with additional spectral image assistance have been proposed and gained great success. Xiong \etal~\cite{xiong2021seeing} introduced a new flash technique for low-light imaging which uses deep-red light for assistance. They utilize the sensitivity of silicon sensors to \SI{660}{nm} deep-red light and design a camera prototype together with a fusion network to reconstruct extra-dim scene images. However, the model suffers from color distortion and is unsuitable for wider application scenarios because deep-red flash is visible to human eyes and can be annoying or even harmful. Instead of using \SI{660}{nm} deep-red light as guidance, Jin \etal~\cite{jin2022RGBNIR} used \SI{850}{nm} near-infrared images to guide the enhancement process. Compared to deep-red images, \SI{850}{nm} NIR images suffer from structural discrepancy from corresponding RGB images under certain circumstances (\eg, shadows and dyes). To overcome this issue, they proposed the Deep Inconsistency Prior (DIP) to adaptively leverage the structure inconsistency to guide the fusion of RGB-NIR. 

\noindent \textbf{NIR-to-RGB Translation.}
NIR-to-RGB Translation aims to colorize a NIR image into an RGB image. Limmer \etal~\cite{limmer2016infrared} first trained a deep multi-scale convolutional neural network that performs direct and integrated transfer between NIR and RGB pixels. 
Su{\'a}rez \etal~\cite{suarez2017infrared} learned each color channel independently for NIR colorization based on the usage of a triplet model to pursue fast convergence and greater similarities. Wang \etal~\cite{Wang_2020_VCIP_Semantic} proposed a multi-task framework that employs additional supervision, such as semantic loss, to aid in the NIR colorization process. To deal with the unpaired data, Nyberg \etal~\cite{nyberg2018unpaired} and Mehri \etal~\cite{mehri2019colorizing} learned the mapping with an unsupervised Generative Adversarial Network (GAN)~\cite{Goodfellow_2014_nips_GAN} based on CycleGAN~\cite{zhu2017unpaired}. However, the outputs of these methods suffer from extreme blurring as well as texture and chrominance mismatching due to the poor associations between outputs and ground truth images. Wu \etal~\cite{wu2020learn} proposed a supervised learning-based method for NIR2RGB video translation, yet the training data were captured in the daytime, and the gap of illumination distribution between artificial LEDs and natural illuminants still exists. 

\noindent \textbf{NIR and RGB Image Fusion.}
Traditional image fusion methods were often based on spatial transformation techniques such as wavelet transform~\cite{lewis2007pixel}, contourlet transform~\cite{da2006nonsubsampled}, and edge-preserving filter-based transform~\cite{ma2017infrared}. Yan \etal~\cite{yan2015infrared} proposed a novel fusion method based on the spectral graph wavelet transform (SGWT) and the bilateral filter. Hu \etal~\cite{hu2017adaptive} used the cumulative distribution of gray levels and entropy to adaptively retain infrared-hot targets and visible textures while fusing infrared and visible videos. Due to the rapid development of deep learning in recent years, many learning-based methods have attracted great attention and gained great success in this field~\cite{Li_2019_TIP_FusionApproach,xu2019learning,xu2020u2fusion}. Li \etal~\cite{Li_2019_TIP_FusionApproach} proposed an end-to-end deep architecture with dense blocks and fusion layers to fuse infrared and visible images in one forward pass. 
DDcGAN \cite{xu2019learning} was proposed with a special dual-discriminator design to generate relatively realistic visible images of different resolutions. 
U2Fusion~\cite{xu2020u2fusion} was proposed to automatically estimate the importance of corresponding source images with adaptive information preservation degrees. 

\noindent \textbf{Hyperspectral Image Datasets.}
Various hyperspectral dataset has been proposed in order to analyze the characteristics of different wavelengths. Arad~\etal~\cite{arad2016sparse} proposed an ICVL dataset that contains hyperspectral data of 201 different scenes. The dataset was taken in sufficient light using a Specim PS Kappa DX4 hyperspectral camera and a rotary stage for spatial scanning, and most of them are captured outdoors. Monno~\etal~\cite{monno2018single} proposed the TokyoTech dataset containing 59-band visible-NIR hyperspectral images from \SI{420}{nm} to \SI{1000}{nm} at \SI{10}{nm} intervals. The images were captured using a monochrome camera and two VariSpec tunable filters, VIS for \SI{420}{nm}-\SI{650}{nm} and SNIR for \SI{650}{nm}-\SI{1000}{nm}, for capturing each hyperspectral image. Liu~\etal~\cite{Liu_2022_CVPR} built a complex imaging system and contributed an IDH dataset containing hyperspectral images from \SI{420}{nm} to \SI{1000}{nm} at \SI{10}{nm} intervals. The UI-3860CP grayscale camera together with the Kurios-XE2 tunable filter is used to record spectral images from \SI{650}{nm} to \SI{1000}{nm}, while the 15S5C camera is used to record the RGB image. In this paper, we contribute a new hyperspectral image dataset to supplement the existing HSI datasets in terms of quality and quantity.

\section{Method}

In this section, we present our Visibility Constrained Spectrum Design (VCSD) method. We first introduce the human vision system in Sec.\ref{sec:visionsys} as the prerequisite of the VCSD model, which will be described in Sec.\ref{sec:MVM}. After that, we introduce the physic-based Imaging Process Simulation (IPS) model in Sec.\ref{sec:IPS}. We then describe our Image Restoration model in Sec.\ref{sec:network}. Finally, we sum up the training procedure in Sec.~\ref{sec:procedure}.

\begin{figure}
\centering
 \subcaptionbox{}{\includegraphics[width = 0.23\textwidth]{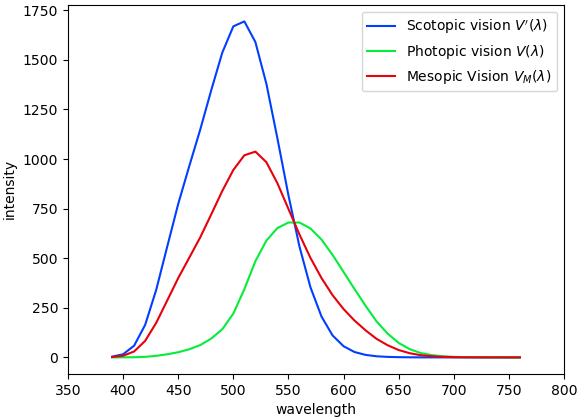}}
 \subcaptionbox{}{\includegraphics[width = 0.23\textwidth]{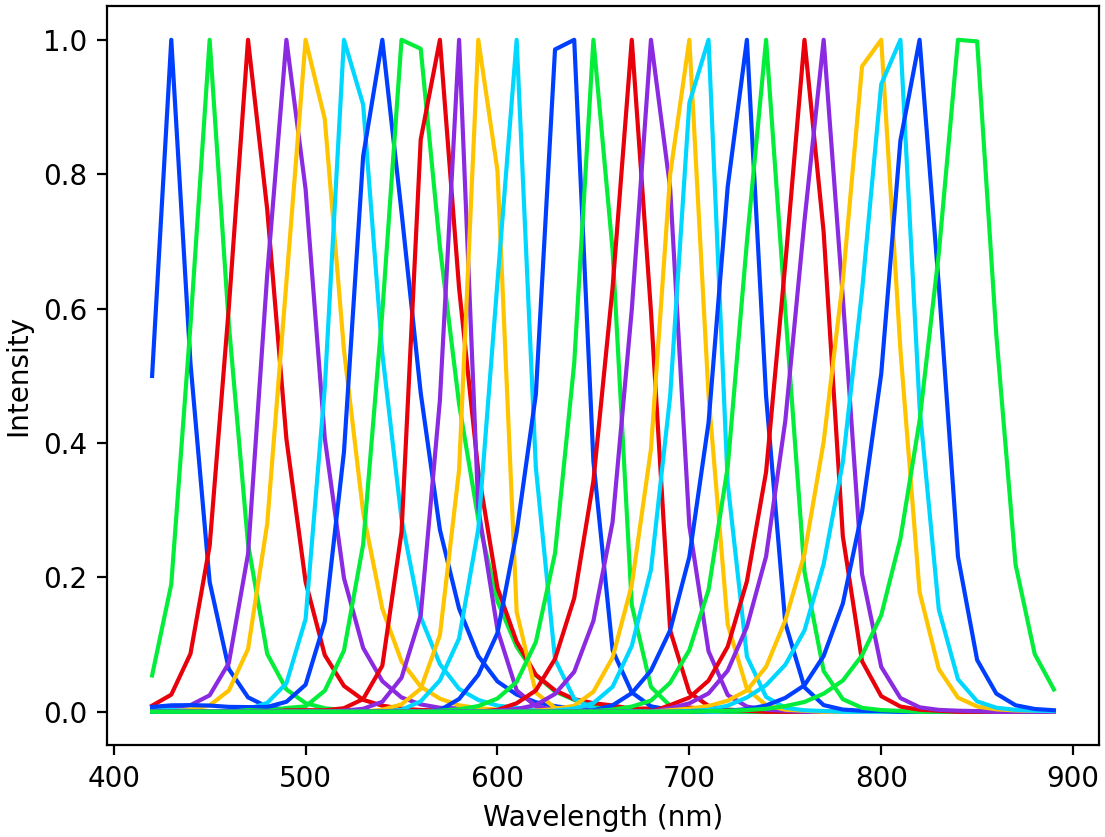}}
\caption{(a) Photopic, scotopic, and mesopic intensity functions. (b) The spectra of 26 narrow band LEDs used in our experiments. }
\label{fig:curves}
\end{figure}

\subsection{Human Vision System}
\label{sec:visionsys}
The human eye is sensitive to wavelengths roughly between \SI{400}{nm} and \SI{700}{nm}. Wavelengths shorter than \SI{400}{nm} or longer than \SI{700}{nm} are almost invisible. For wavelengths between \SI{400}{nm} and \SI{700}{nm}, human eyes behave differently in high or low light conditions. In relatively high light conditions, the vision is mainly relevant to the center of the retina whose maximum sensitivity is at \SI{555}{nm} (in the green region). This type of vision is called photopic vision. While in extremely low light conditions for human eyes, which is our case, the vision is done by the peripheral region of the retina whose maximum sensitivity is at \SI{507}{nm} (in the blue-green region). This type of vision is called scotopic vision. At intermediate light levels, both rods and cones are active, which is called mesopic vision. As shown in Fig.\ref{fig:curves}(a), given the photopic and scotopic luminosity functions as $V(\lambda)$ and $V'(\lambda)$, the mesopic luminosity function $V_M(\lambda)$ can be approximated as: 
\begin{align}
V_M(\lambda)=(1-x)V'(\lambda)+xV(\lambda),
\end{align}
where $x$ is determined by photopic illuminance and the composition of light source~\cite{crawford1949scotopic,wald1945human,eloholma2006new,xiong2021seeing}. 

\subsection{Visibility Constrained Spectrum Design}
\label{sec:MVM}
To consider different wavelengths, we propose to find an optimal LED spectral multiplexing based on $K$ LED bases of different wavelengths:
\begin{align}
\label{equ:multiplex}
\Phi(\lambda) = \sum^{K-1}_{k=0} \sigma^k \cdot \Phi^k(\lambda),
\end{align}
where $\boldsymbol{\sigma} = [\sigma^0, \sigma^1, ..., \sigma^{K-1}], \sigma^k \in (0,1)$ is the parameter that determine the weight of corresponding LED base $\Phi^k(\lambda)$, and can be optimized during the training process.

To consider the human vision system, the key issue is to find a way to quantify the visibility of certain LED spectral curves. Given the scotopic intensity functions $V'(\lambda)$ and the LED spectral curve $\Phi(\lambda)$, The perceived power $P_m$ of light by human naked eyes is proportional to the inner product of $V'(\lambda)$ with $\Phi(\lambda)$~\cite{xiong2021seeing}:
\begin{align}
P_m \propto \int V'(\lambda)\Phi(\lambda)d\lambda = \Psi.
\end{align}
Since the relationship is proportional, not equal, it is hard to get the real value of $P_m$ given certain $V'(\lambda)$ and $\Phi(\lambda)$. However, it is indeed possible to \textit{determine a threshold $\hat{\Psi}$ through user studies, which represents 'just' invisible to the human eye.}

Therefore, given the threshold $\hat{\Psi}$ and a mutiplexed LED spectrum $\Phi(\lambda)=\sum^{K-1}_{k=0} \sigma^k \cdot \Phi^k(\lambda)$ that is visible to human eyes (\ie, $\int V'(\lambda)\Phi(\lambda)d\lambda > \hat{\Psi}$), we calculate a scale factor $\xi \in (0, 1)$ so that when $\hat{\sigma}^k = \xi \cdot \sigma^k, (k = 0,1,...,K-1)$, $\hat{\Phi}(\lambda)=\sum^{K-1}_{k=0} \hat{\sigma}^k \cdot \Phi^k(\lambda)$ becomes just invisible to human eyes, \ie, the perceived power of light by human scotopic vision equals to $\hat{\Psi}$:
\begin{align}
\int V'(\lambda)\hat{\Phi}(\lambda)d\lambda = \ &\hat{\Psi}. \\
\therefore \ \xi = \frac{\hat{\Psi}}{\int V'(\lambda)\Phi(\lambda)d\lambda}&. \\
\Rightarrow \ \hat{\sigma}^k = \frac{\hat{\Psi}}{\int V'(\lambda)\Phi(\lambda)d\lambda + \epsilon} \cdot \sigma^k, k &= 0,...,K-1,
\label{equ:xi}
\end{align}
where $\epsilon$ is a small constant to avoid numerical issues.
Therefore, given an LED spectrum $\Phi(\lambda)$ that is visible to human eyes, we scale it by $\xi$ calculated from Eq.\ref{equ:xi} to make it just invisible to human naked eyes. Note that for LED spectrum that is already `invisible' to human naked eyes (\ie, $\int V'(\lambda)\Phi(\lambda)d\lambda < \hat{\Psi}$), we just let $\xi=1$ since there is no need to adjust the spectral curve intensity of these LEDs.

\subsection{Imaging Process Simulation Model}
\label{sec:IPS}
In computational photography, the formation of images depends on three factors: the reflectance spectrum $\mathcal{T}$, the illumination spectrum $\Phi$, and the camera spectral sensitivity $\mathcal{C}$. Given these three factors, the process of acquiring light intensity for each pixel can be formulated as:
\begin{align}
\label{equ:acqu}
    \mathcal{I}_{c,i,j} = \int \mathcal{T}_{i,j}(\lambda) \cdot \Phi(\lambda) \cdot \mathcal{C}_{c}(\lambda) \,d\lambda,
\end{align}
where $c \in \{R,G,B\}$ represents the color channel. $i \in \{1, 2, ..., W\}, j \in \{1, 2, ..., H\}$, $W$ and $H$ is the width and height of the image. $\mathcal{I}_{c,i,j}$ denotes the RGB intensity in channel $c$ at position $(i, j)$. $\mathcal{T}_{i,j}(\lambda)$ is the reflectance spectrum in position $(i, j)$. $\Phi(\lambda)$ is the LED spectrum. $\mathcal{C}_{c}(\lambda)$ is the camera spectral sensitivity in channel $c$. The image can be obtained according to $\mathcal{I}_{c,i,j}$ in each position.

Based on this physical process, we design an Imaging Process Simulation (IPS) model to generate assistance images. The IPS module takes assistance spectral curve $\Phi \in \mathbb{R}^{L}$, camera spectral sensitivity $\mathcal{C} \in \mathbb{R}^{3 \times L}$, and reflectance spectrogram $\mathcal{T} \in \mathbb{R}^{L \times W \times H}$ as inputs, and outputs the synthesized assistance images $\mathcal{I} \in \mathbb{R}^{3 \times W \times H}$. We set the range of camera spectral sensitivity to \SI{420}{nm}-\SI{890}{nm}, with a \SI{10}{nm} interval, so the process can be formulated as:
\begin{align}
\label{equ:2}
    \mathcal{I}_{c,i,j} = \sum_{n=0}^{47} \mathcal{T}_{i,j}(n) \cdot \Phi(n) \cdot \mathcal{C}_{c}(n),
\end{align}
where $n = 0, 1, ..., 47$ represents $48$ different wavelengths covered by the camera spectral sensitivity. Note that this process is fully differentiable under Equ.\ref{equ:2}, allowing us to optimize the parameter $\boldsymbol{\sigma}$ that determines the weight of each LED base.

\noindent \textbf{Noise Simulation.}
In low-light environments, assistance images are usually free of obvious noise interference due to enough illumination provided by the LEDs, but noise still exists under these conditions. Also, according to our Visibility Model, the intensity of LED may become very small in order to become invisible to the human naked eyes, which makes the assistance images suffer from obvious noise interference. Since Equ.~\ref{equ:2} can't model the real camera noise widely existing during the image formulation, we additionally introduce a noise model to consider the noise effects for assistance images.

Poisson Distribution has been widely considered to model the noise distribution~\cite{wei2020physics,zhang2021rethinking,feng2022learnability}. Here we choose to combine Poisson Distribution with noise sampling from a real camera sensor to realize our noise model:
\begin{align}
    \hat{\mathcal{I}} = \kappa \cdot \mathcal{P}(\frac{\mathcal{I} \cdot \xi }{\kappa})+\mathcal{N},
\end{align}
where $\kappa$ is the gain of the target camera, $\xi$ is the scale-down factor calculated in the Visibility Constrained Spectrum Model, and $\mathcal{N}$ is the real noise pattern sampled from the target camera. 

\begin{algorithm}[t]
\caption{Training}
\label{algorithm}
\hspace*{0.02in} {\bf Input: }
Visibility Threshold $\hat{\Psi}$, LED bases $[\Phi^k]_{k=0}^{K-1}$, Camera gain $\kappa$, Camera Spectral Sensitivity $\mathcal{C}$, Hyper-spectral Dataset $\mathcal{D}_{tr}$, and hyperparameters $\boldsymbol{\sigma}, \boldsymbol{\theta}$. \\
\hspace*{0.02in} {\bf Output:}
Optimal wide-band spectral curve.
\begin{algorithmic}[1]

\State Initialize $\boldsymbol{\sigma} \leftarrow \boldsymbol{\sigma}_t = [\sigma^0_{t}, \sigma^1_{t}, ..., \sigma^{K-1}_{t}]$.
\While {not converged}
    \State Randomly sample $\mathcal{T}_t$, $\mathcal{Y}_t$, $\mathcal{N}_t^{vis}$, $\mathcal{N}_t^{nir}$ from $\mathcal{D}_{tr}$.
    
    \State $\Phi_t(\lambda) \leftarrow \sum^{K-1}_{k=0} \sigma^k_{t} \cdot \Phi^k(\lambda)$

    \State $\mathcal{T}_t^{vis} \leftarrow \mathcal{T}_t^{420:700nm}$, $\Phi_t^{vis} \leftarrow \Phi_t^{420:700nm}$

    \State $\xi_t \leftarrow \operatorname{min}(\frac{\hat{\Psi}}{\int V'(\lambda)\Phi_t(\lambda)d\lambda + \epsilon}, 1)$

    \For{$k \leftarrow 0$ to $K-1$}
    \If{$\int V'(\lambda)\Phi^k(\lambda)d\lambda>0$}
    \State $\hat{\sigma}^k_t \leftarrow \xi_t \cdot \sigma^k_t$
    \Else
    \State $\hat{\sigma}^k_t \leftarrow \sigma^k_t$
    \EndIf
    \EndFor

    \State $\hat{\Phi}_t(\lambda) = \sum^{K-1}_{k=0} \hat{\sigma}^k_t \cdot \Phi^k(\lambda)$
    
    \State $\hat{\Phi}_t^{vis} \leftarrow \hat{\Phi}_t^{420:700nm}$
    
    \State $\xi^{vis}_t \leftarrow \frac{\int \hat{\Phi}^{vis}_t(\lambda)d\lambda}{\int\Phi^{vis}_t(\lambda)d\lambda + \epsilon}, \ \xi^{nir}_t \leftarrow \frac{\int \hat{\Phi}_t(\lambda)d\lambda}{\int\Phi_t(\lambda)d\lambda + \epsilon}$

    
    \State $\mathcal{I}^{vis}_{c,i,j,t} \leftarrow \int \limits \mathcal{T}^{vis}_{i,j,t}(\lambda) \cdot \hat{\Phi}^{vis}_t(\lambda) \cdot \mathcal{C}_{c}(\lambda) d\lambda$
    \State $\mathcal{I}^{nir}_{c,i,j,t} \leftarrow \int \limits \mathcal{T}_{i,j,t}(\lambda) \cdot \hat{\Phi}_t(\lambda) \cdot \mathcal{C}_{c}(\lambda) d\lambda$
    
    
    \State $\hat{\mathcal{I}}^{vis}_t \leftarrow \kappa \cdot \mathcal{P}(\frac{\mathcal{I}^{vis}_t \cdot \xi^{vis}_t }{\kappa})+\mathcal{N}_t^{vis}$
    \State $\hat{\mathcal{I}}^{nir}_t \leftarrow \kappa \cdot \mathcal{P}(\frac{\mathcal{I}^{nir}_t \cdot \xi^{nir}_t }{\kappa})+\mathcal{N}_t^{nir}$
    
    \State $\mathcal{X}_t \leftarrow G(\hat{\mathcal{I}}^{vis}_t, \hat{\mathcal{I}}^{nir}_t; \boldsymbol{\theta})$
    \State Take gradient descent step on
    \State \ \ \ \ \ \ \ \ \ \ \ \ \ $\nabla_{\boldsymbol{\theta}, \boldsymbol{\sigma}} \mathcal{L}(\mathcal{X}_t, \mathcal{Y}_t)$
\EndWhile
\State \Return Optimized spectral curve $\boldsymbol{\sigma}^*$
\end{algorithmic}
\end{algorithm}

\subsection{Image Restoration Model}
\label{sec:network}
\noindent \textbf{Network Architecture.} Our fusion network takes a VIS image $\hat{\mathcal{I}}^{vis}$ and an NIR-VIS image $\hat{\mathcal{I}}^{nir}$ as input and generates the result $\mathcal{X}$. We choose the same UNet~\cite{ronneberger2015u} structure as ~\cite{Liu_2022_CVPR} during the curve design process, except for the number of input channels, which is 3 in~\cite{Liu_2022_CVPR}, and 6 in our work.

\noindent \textbf{Loss Function.} The perceptual loss~\cite{johnson2016perceptual} has been widely used in image reconstruction tasks due to its ability to recover details and preclude over-smooth results compared to pixel-wise losses:
\begin{align} 
\mathcal{L} = \sum_{i=1}^I \left\|\mathcal{\psi}_{i}(\mathcal{X}) - \mathcal{\psi}_{i}(\mathcal{Y})\right\|_{1},
\end{align}
where $\mathcal{X}$ is the output of $G$, and $\mathcal{Y}$ is the corresponding ground truth. $\mathcal{\psi}_i$ denotes the activation map at the $i$-th layer of the pre-trained VGG-19 network \cite{simonyan2014very}. Particularly, we chose 5 layers including $relu_{1-1}$, $relu_{2-1}$, $relu_{3-1}$, $relu_{4-1}$, and $relu_{5-1}$ from the VGG-19 network.

\begin{figure}[t]
\centering
\setlength{\abovecaptionskip}{1mm}
\includegraphics[width=\linewidth]{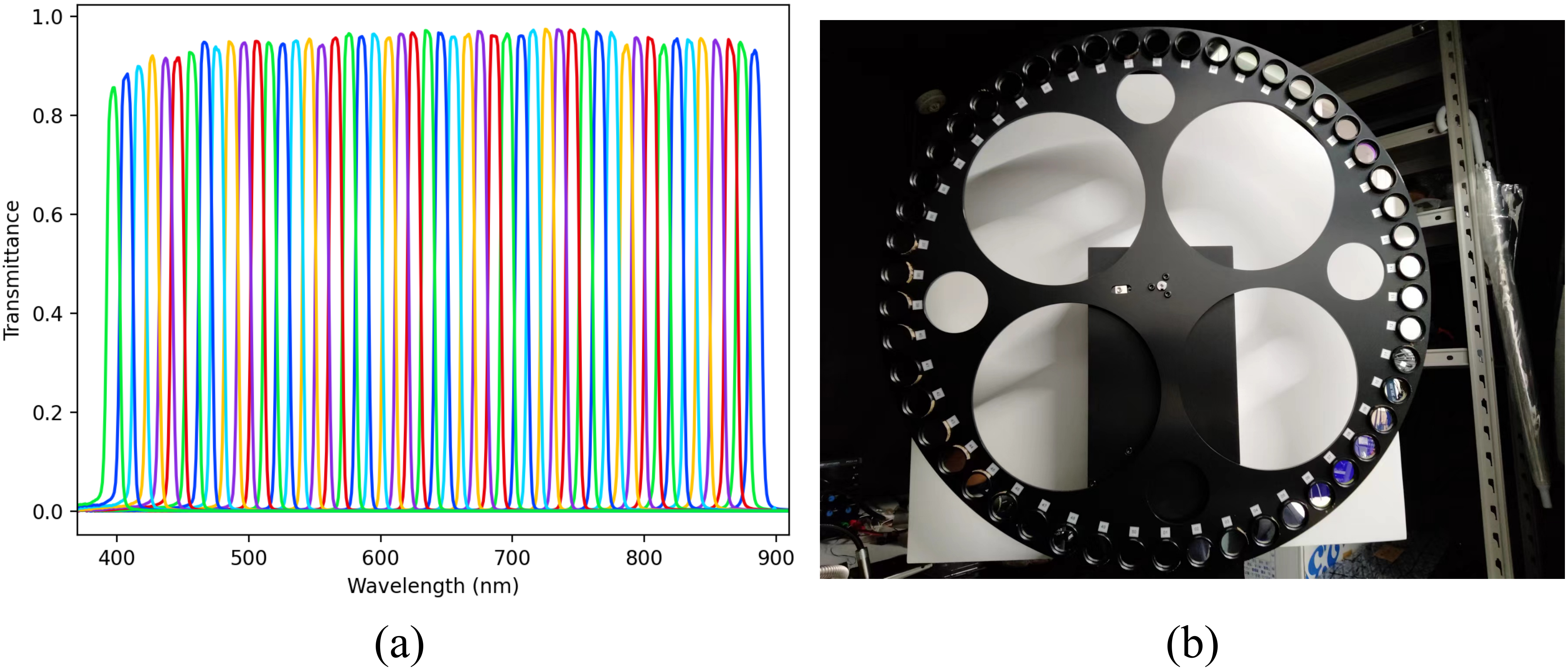}
\caption{(a) The transmittance curves of 50 band pass filters. (b) Camera system to capture the dataset.}
\label{fig:camera}
\vspace{-2mm}
\end{figure}

\begin{table}[t]
\caption{Comparison between different Hyperspectral Datasets.}
\resizebox{\linewidth}{!}{
\begin{tabular}{lcccc}
\toprule
Datasets                    & ICVL                       & TokyoTech                  & IDH                        & Ours                       \\ \midrule
Resolution  & 1392$\times$1300         & 512$\times$512           & 256$\times$256           &    1936 $\times$ 1096                \\
Scenes              & 201                        & 16                         & 112                        &    74                        \\
Range/nm    & 400-1000        & 420-1000        & 650-1000        & 400-890        \\
Interval/nm & 1.25                     & 10                       & 10                       & 10                       \\
\bottomrule
\end{tabular}
}
\label{tab:hsi}
\vspace{-4mm}
\end{table}

\newlength{\hs}
\setlength{\hs}{0.002\textwidth}
\begin{figure*}[!t]
\setlength{\abovecaptionskip}{2mm}
\includegraphics[width=\linewidth]{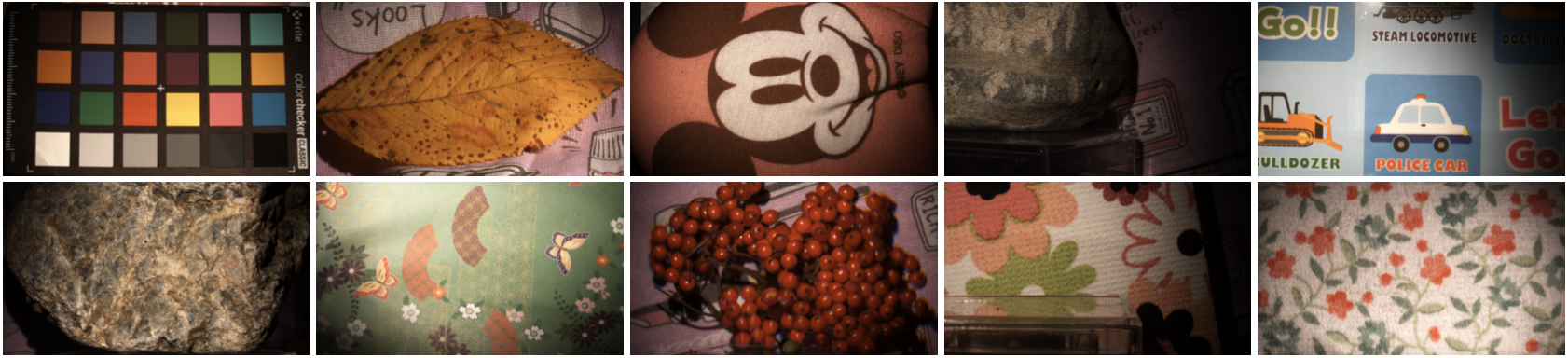}
\caption{Data samples from our proposed dataset. The images are synthesized using the spectrum of white-light LED.}
\label{fig:datasample}
\vspace{-4mm}
\end{figure*}

\subsection{Training Procedure}
\label{sec:procedure}
Algorithm \ref{algorithm} displays the complete training procedure. We first initialize the parameter $\boldsymbol{\sigma}$ that determines the weight of each LED base, which will be optimized by the gradient. In each iteration, scale factor $\xi_t$ is first calculated and used to make the designed spectrum $\Phi_t$ invisible. Note that since there exist several NIR LED bases whose spectrum curves have no intersection with scotopic intensity functions (\ie, $\int V'(\lambda)\Phi^k(\lambda)d\lambda=0$), we may not want to scale down the corresponding coefficient $\sigma^k_t$ since it provides no improvements for visibility but causes information loss. As a result, we choose to only scale down the coefficients of LED bases whose spectrum have intersections with scotopic intensity functions, and obtain the new curve $\hat{\Phi}_t(\lambda)$ that also fulfills the visibility limitations:
\begin{align}
\xi_t = \operatorname{min}(\frac{\hat{\Psi}}{\int V'(\lambda)\Phi_t(\lambda)d\lambda + \epsilon}, 1)
\end{align}
\vspace{-5mm}
\begin{align}
\hat{\sigma}^k_t = \left\{ 
    \begin{aligned} 
    &\xi_t \cdot \sigma^k_t, \int V'(\lambda)\Phi^k(\lambda)d\lambda>0 \\
    &\sigma^k_t, \int V'(\lambda)\Phi^k(\lambda)d\lambda=0
    \end{aligned} 
\right.
\end{align}
\vspace{-5mm}
\begin{align}
\hat{\Phi}_t(\lambda) = \sum^{K-1}_{k=0} \hat{\sigma}^k_t \cdot \Phi^k(\lambda).
\end{align}

We then calculate the scale-down factor for VIS images and NIR images as:
\begin{align}
\xi^{vis}_t &= \frac{\int \hat{\Phi}^{vis}_t(\lambda)d\lambda}{\int\Phi^{vis}_t(\lambda)d\lambda + \epsilon}, \ \ \xi^{nir}_t = \frac{\int \hat{\Phi}_t(\lambda)d\lambda}{\int\Phi_t(\lambda)d\lambda + \epsilon},
\end{align}
where $\Phi^{vis}_t$ and $\hat{\Phi}_t^{vis}$ are the \SI{420}{nm}-\SI{700}{nm} part of $\Phi_t$ and $\hat{\Phi}_t$, respectively.
Based on $\hat{\Phi}_t(\lambda)$, input image $\hat{\mathcal{I}}^{nir}_t$ and $\hat{\mathcal{I}}^{vis}_t$ are simulated via the physic-based IPS module which considers noise effects related to $\xi^{vis}_t$ and $\xi^{nir}_t$. After that, the output $\mathcal{X}_t$ is obtained through our image enhancement network $G$ which takes auxiliary images $\hat{\mathcal{I}}^{nir}_t$ and $\hat{\mathcal{I}}^{vis}_t$ as input. Finally, gradient descent steps are taken based on the loss function $\mathcal{L}$.

\subsection{Implementation Details}
We implement the training part of our model with Pytorch~\cite{paszke2019pytorch}. During the spectrum optimization process, we set the batch size to 16 and the learning rate to 1e-3. The total training iteration is 50,000, and the learning rate is multiplied by 0.1 every 20,000 iterations. We use Adam optimizer~\cite{kingma2014adam} with $\beta_1 = 0.5, \beta_2 = 0.999$, and randomly crop the input images to 256$\times$256. We set the number $K$ of LEDs to 26, covering the wide-band VIS-NIR range from \SI{420}{nm} to \SI{890}{nm}. The spectrum of these LED bases is shown in Fig.~\ref{fig:curves}(b). We use the camera GS3-U3-15S5C for both image synthesis and real image capture. We choose a normalized \SI{660}{nm} LED spectrum to obtain the visibility threshold $\hat{\Psi}=10$ for our main experiment, following the claim in~\cite{xiong2021seeing}. We also further discuss the effect of different visibility thresholds on our model in the experiment part.

After obtaining the optimal spectrum, we train an image restoration network using synthesized input images. To train the restoration network on our proposed dataset, we set the batch size to 16 and the learning rate to 1e-4. The total training iteration is 10,000 iterations. We use Adam optimizer~\cite{kingma2014adam} with $\beta_1 = 0.5, \beta_2 = 0.999$, and randomly crop the input images to 256$\times$256. To train the restoration model on TokyoTech~\cite{monno2018single} dataset, we set the batch size to 8 and keep the rest of the settings the same as training on our dataset.

\newlength{\hw}
\setlength{\hw}{0.002\textwidth}
\begin{figure*}[!t]
	\centering
	\renewcommand\arraystretch{0.6}
	\begin{tabular}
		{   
			@{\hspace{\hw}}c@{\hspace{\hw}}
			@{\hspace{\hw}}c@{\hspace{\hw}}
			@{\hspace{\hw}}c@{\hspace{\hw}}
			@{\hspace{\hw}}c@{\hspace{\hw}}
			@{\hspace{\hw}}c@{\hspace{\hw}}
		}
		OptNIR~\cite{Liu_2022_CVPR} & *VIS+\SI{850}{nm} & *VIS+\SI{660}{nm} & *VIS+*NIR & GT \\
		\noalign{\vspace{1pt}}
		\includegraphics[width=0.195\textwidth]{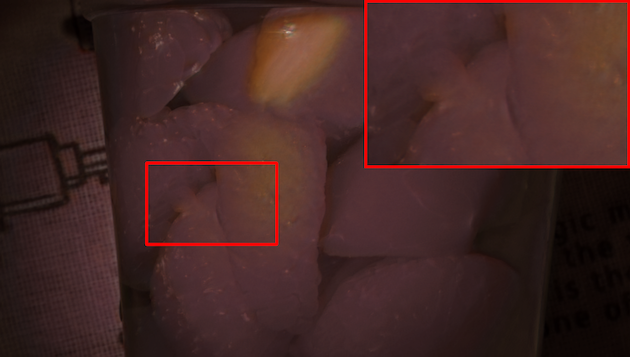} &
		\includegraphics[width=0.195\textwidth]{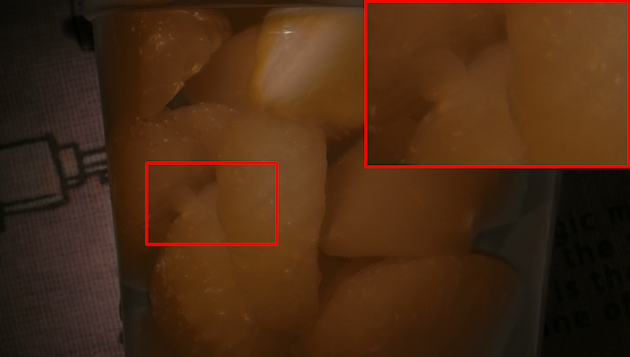} &
		\includegraphics[width=0.195\textwidth]{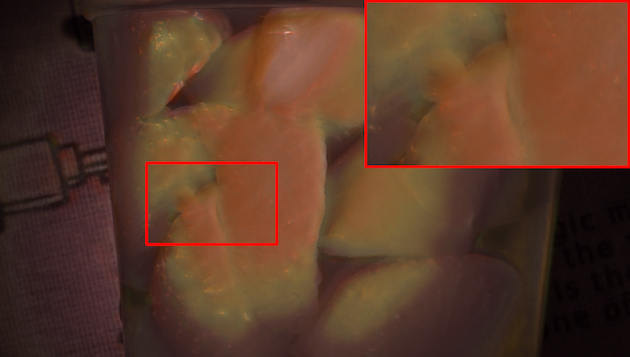} &
		\includegraphics[width=0.195\textwidth]{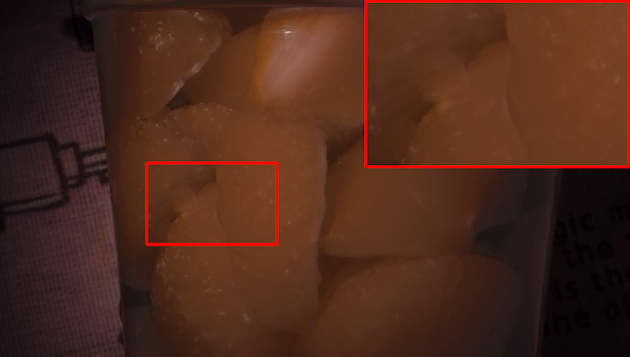} &
		\includegraphics[width=0.195\textwidth]{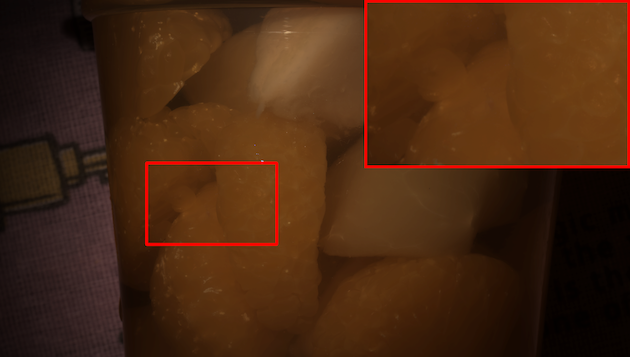} \\
		\includegraphics[width=0.195\textwidth]{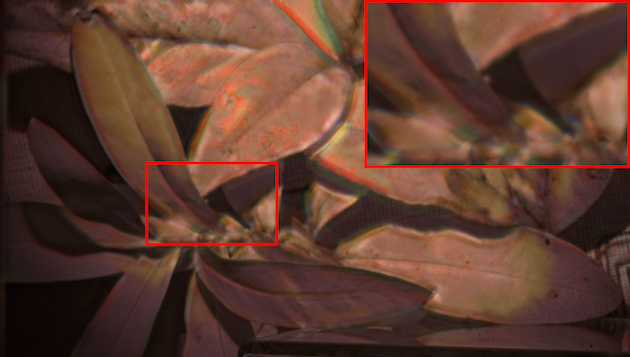} &
		\includegraphics[width=0.195\textwidth]{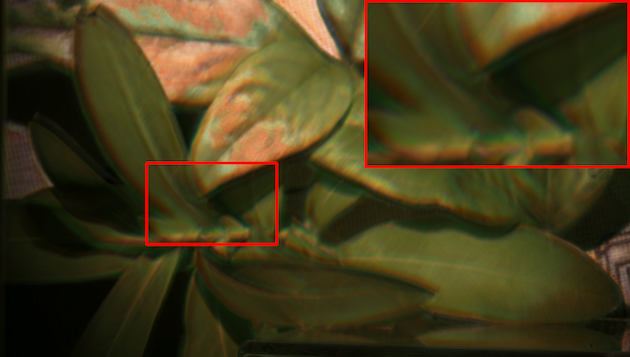} &
		\includegraphics[width=0.195\textwidth]{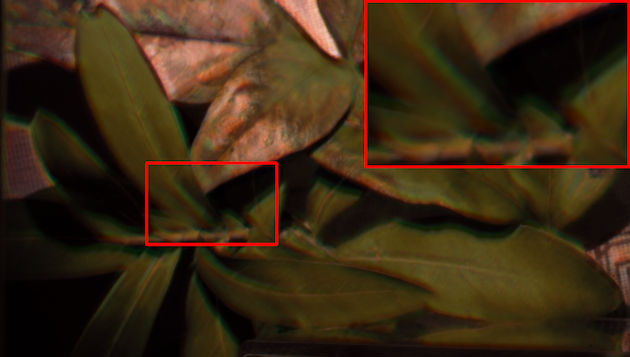} &
		\includegraphics[width=0.195\textwidth]{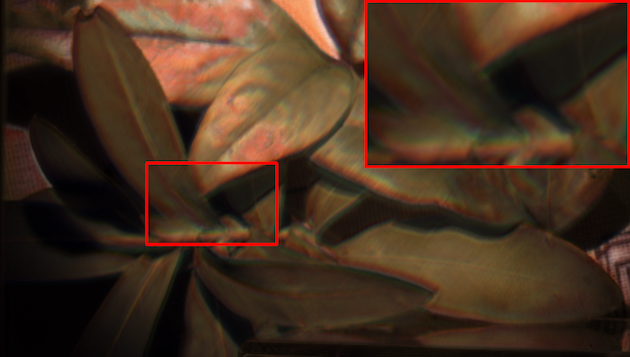} &
		\includegraphics[width=0.195\textwidth]{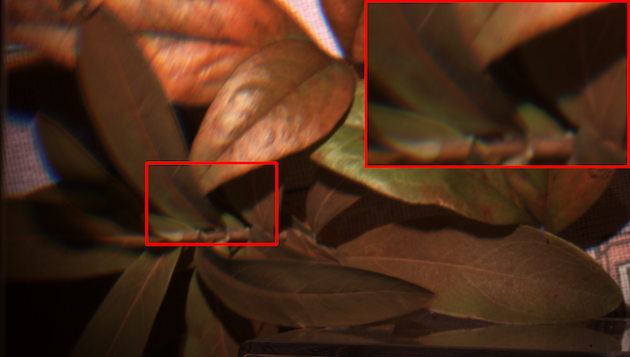} \\

    	\hdashline[3pt/2pt]
    	\noalign{\vspace{2pt}}
  
		\includegraphics[width=0.195\textwidth]{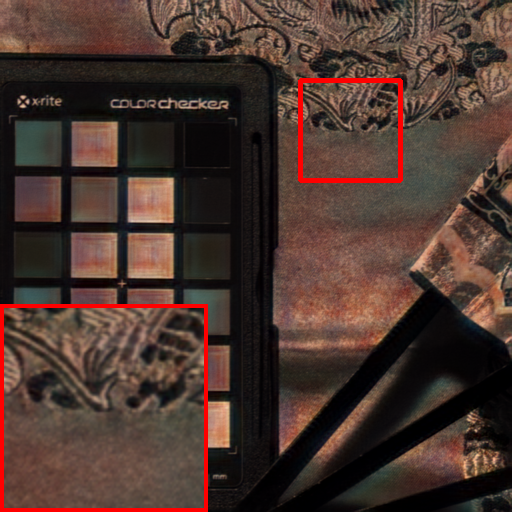} &
		\includegraphics[width=0.195\textwidth]{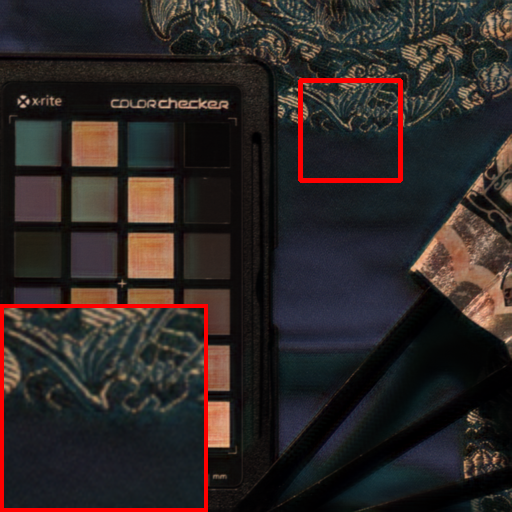} &
		\includegraphics[width=0.195\textwidth]{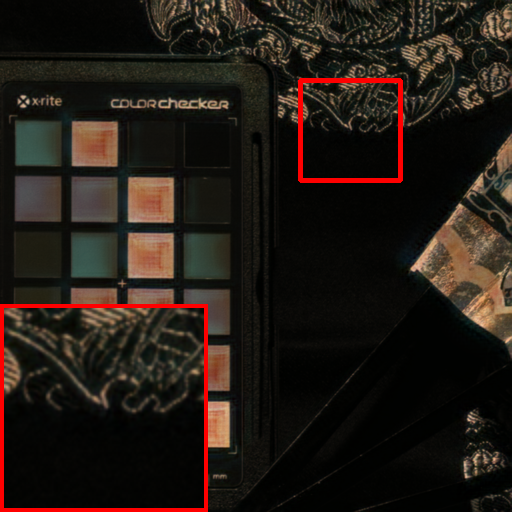} &
		\includegraphics[width=0.195\textwidth]{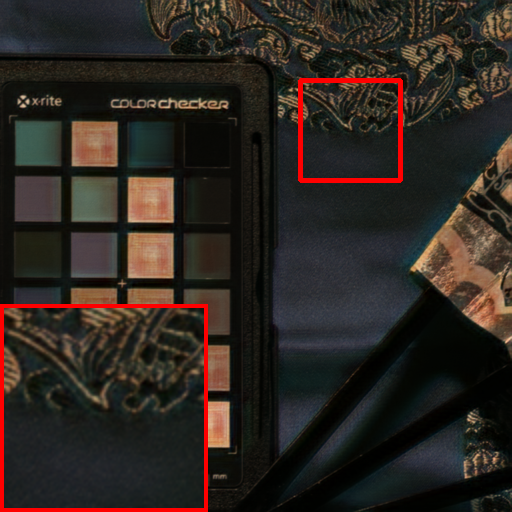} &
		\includegraphics[width=0.195\textwidth]{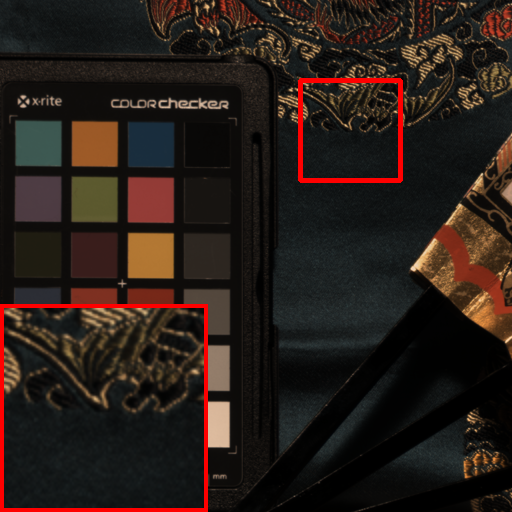} \\

	\end{tabular}
 \setlength{\abovecaptionskip}{1mm}
	\caption{Qualitative results on our dataset (above dash) and the TokyoTech dataset~\cite{monno2018single} (below dash). Please zoom in for a clear view.}
	\label{fig:qualitative}
	\vspace{-3mm}
\end{figure*}

\section{Experiments}

\subsection{Settings}
\noindent \textbf{Datasets.} To train our model, we need datasets that contain hyperspectral images of multiple scenes that cover from VIS range to NIR range.
Up to now, some hyperspectral datasets have been proposed, including ICVL~\cite{arad2016sparse}, TokyoTech~\cite{monno2018single}, and IDH~\cite{Liu_2022_CVPR}. ICVL contains hyperspectral data of 201 scenes, with a spatial resolution of 1392 $\times$ 1300 and 519 spectral bands (\SI{400}{nm} to \SI{1000}{nm} at roughly \SI{1.25}{nm} increments). TokyoTech contains 59-band hyperspectral images from \SI{420}{nm} to \SI{1000}{nm} at \SI{10}{nm} intervals. The image resolution is 512 $\times$ 512, and only 16 scenes are publicly available. IDH dataset contains a total of 112 hyperspectral images from \SI{650}{nm} to \SI{1000}{nm} at \SI{10}{nm} intervals, and the spatial resolution is only 256 $\times$ 256. In this paper, we contribute a new hyperspectral image dataset to supplement existing HSI datasets in terms of quality and quantity. The wavelength covers from \SI{400}{nm} to \SI{890}{nm} with \SI{10}{nm} intervals. There are 74 scenes in the dataset, with a resolution of 1936 $\times$ 1096. The ground truth VIS image is synthesized via the white-light LED spectral curve. A comparison between different hyperspectral image datasets is shown in Tab.~\ref{tab:hsi}.

We test the effect of our designed spectrum on two datasets: our proposed dataset and the TokyoTech dataset. To test the result, we first obtain the two auxiliary images according to the designed curve, then train an image enhancement network on the training set. We then test the effect of this network on the test set. The synthetic dataset comes from our collected hyperspectral dataset. Two auxiliary images and the corresponding ground truth are obtained through the physic-based imaging process described in Equ.~\ref{equ:2}. The experimental results will be introduced in Sec.~\ref{sec:synthetic}.

\noindent \textbf{Methods.}
Based on existing popular solutions for low-light imaging, we conduct experiments on three different settings of our method: 1) *VIS+*NIR: Optimal spectrum design for both VIS and NIR (our main method), 2) *VIS+\SI{850}{nm}:  Optimal spectrum design for VIS + fixed \SI{850}{nm} Auxiliary Illumination (based on~\cite{jin2022RGBNIR}), and 3) *VIS+\SI{660}{nm}: Optimal spectrum design for VIS + fixed \SI{660}{nm} Auxiliary Illumination (based on~\cite{xiong2021seeing}). All the VIS images are synthesized based on the \SI{420}{nm}-\SI{700}{nm} spectrum. We then compare these three settings with \cite{Liu_2022_CVPR}, which achieve superior results than traditional RGB-independent methods by retrieving the optimal NIR spectrum whose wavelength is larger than \SI{700}{nm}.

\subsection{Main Results}
\label{sec:synthetic}

\begin{table}[t]
\caption{Quantitative results on different synthetic datasets. The best results are in \bestscore{red} whereas the second best are in \secondscore{blue}.}
\resizebox{\linewidth}{!}{
\begin{tabular}{clccc}
\toprule
Datasets                    & Methods    & SSIM $\uparrow$                                  & PSNR $\uparrow$                                 & LPIPS $\downarrow$                                 \\
\midrule
& OptNIR~\cite{Liu_2022_CVPR}       & 0.7688                                 & 22.67                                 & 0.1590                                 \\
& *VIS+\SI{850}{nm} & 0.8305                                 & \secondscore{24.07} & 0.1276                                 \\
& *VIS+\SI{660}{nm} & \secondscore{0.8316} & 23.72                                 & \secondscore{0.1232} \\
\multirow{-4}{*}{Ours}      & *VIS+*NIR  & \bestscore{0.8383} & \bestscore{24.12} & \bestscore{0.1129} \\
\midrule
& OptNIR~\cite{Liu_2022_CVPR}       & 0.7197                                 & 19.65                                 & 0.1841                                 \\
& *VIS+\SI{850}{nm} & \secondscore{0.7902} & \secondscore{21.78} & \bestscore{0.1365} \\
& *VIS+\SI{660}{nm} & 0.7628                                 & 21.45                                 & 0.1383                                 \\
\multirow{-4}{*}{TokyoTech} & *VIS+*NIR  & \bestscore{0.7938} & \bestscore{22.08} & \secondscore{0.1378} \\
\bottomrule
\end{tabular}
}
\label{tab:quantitative}
\end{table}

In this section, we test different methods on synthetic datasets including our proposed dataset and TokyoTech~\cite{monno2018single}. Quantitative results on our dataset and the TokyoTech dataset are reported in Tab.~\ref{tab:quantitative}. During the evaluation process, SSIM (Structural Similarity), Peak Signal-to-Noise Ratio (PSNR), and Learned Perceptual Image Patch Similarity (LPIPS) are utilized to quantify the difference between restored images and ground truth. Visual results on our dataset and the TokyoTech dataset are shown in Fig.~\ref{fig:qualitative}. We can see that three different settings of our method all perform significantly better than \cite{Liu_2022_CVPR} in three metrics. Instead of empirically picking up the NIR range beyond \SI{700}{nm}, we find the theoretically optimal curve under the visibility constraint $\hat{\Psi}$ by incorporating the human vision intensity function into the optimization process. As a result, we significantly robustify the task and achieve superior results on two datasets. Furthermore, if we use the combination of optimal VIS + empirically fixed NIR spectrum (\eg, \SI{660}{nm}, \SI{850}{nm}), the results may get worse since these combinations are not optimal under the visibility constraint, which also proves the effectiveness of our visibility constrained spectrum design model.

\begin{figure}[t]
\setlength{\abovecaptionskip}{-1mm}
\includegraphics[width=\linewidth]{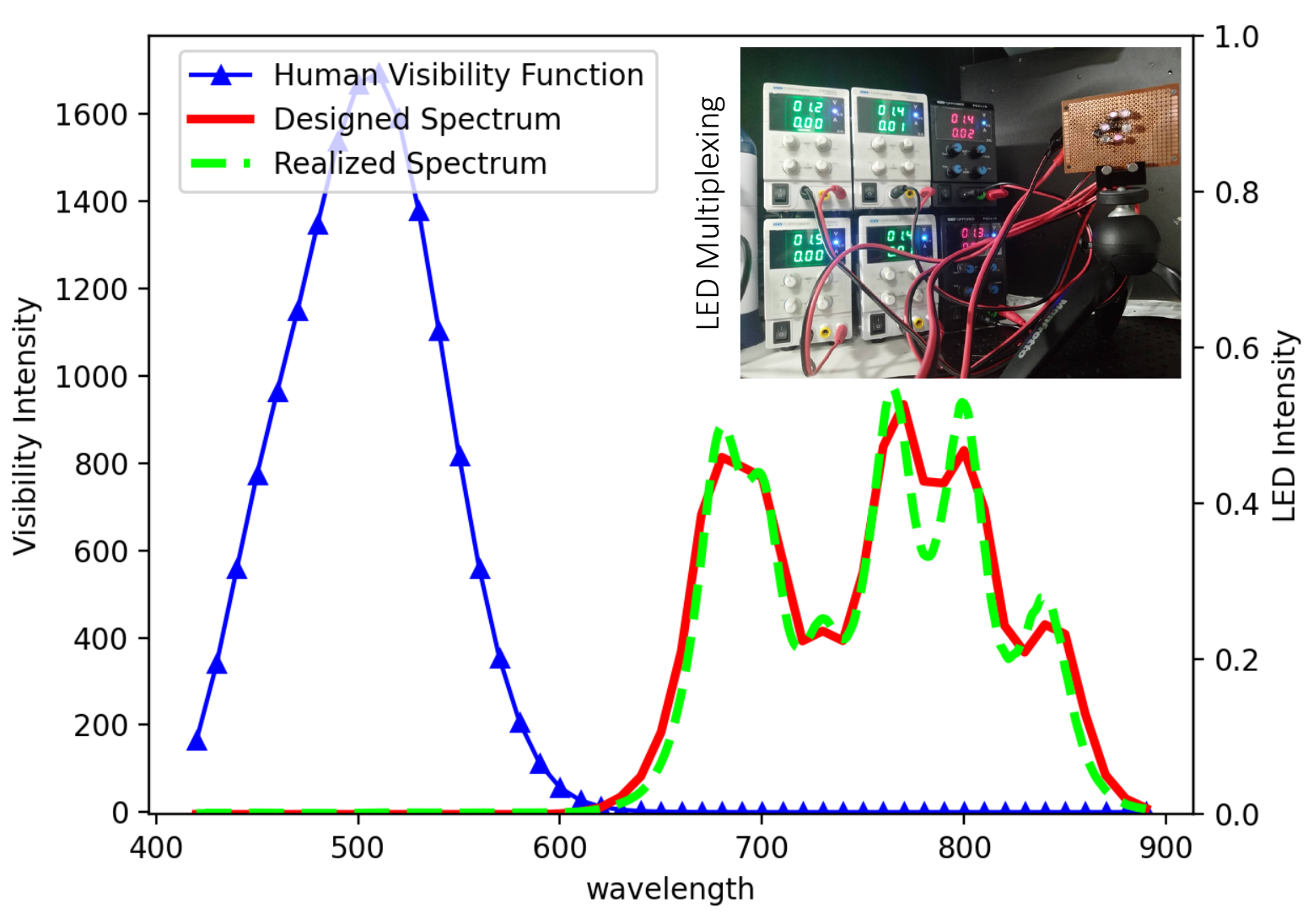}
\caption{Realization for our designed LED spectrum.}
\label{fig:realize}
\vspace{-3mm}
\end{figure}

\noindent \textbf{Designed Curve.} The red line in Fig.~\ref{fig:realize} shows the optimal curve for $\hat{\Psi}$=10 desinged by our model. To further demonstrate the practicality of our model, we implemented the designed spectrum under $\hat{\Psi}$=10 by properly multiplexing different LED bases. As shown in Fig.~\ref{fig:realize}, we approximately fit the designed curve by controlling the voltage of six LED bases, demonstrating the practicality of our designed curve.

\subsection{Impact of Visibility Threshold}

\begin{figure*}[t]
\includegraphics[width=\linewidth]{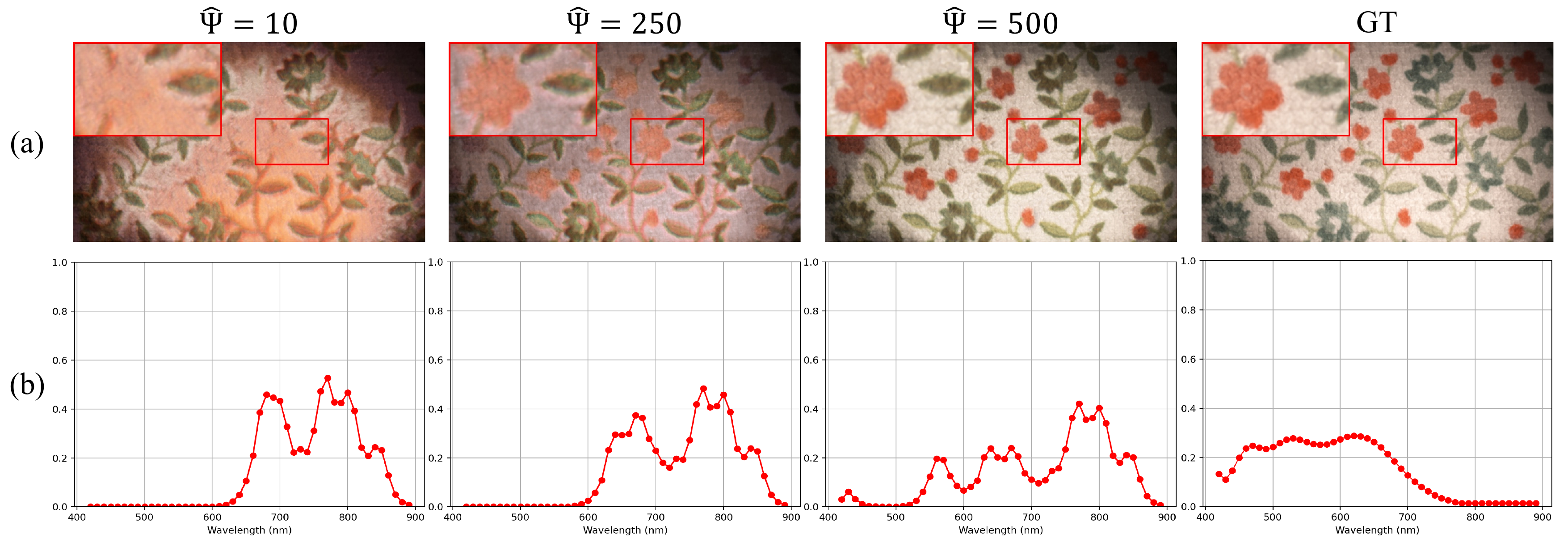}
\setlength{\abovecaptionskip}{-2mm}
\caption{Qualitative results for the impact of different visibility thresholds. (a) Visual results on our proposed dataset, (b) designed curve under the corresponding threshold $\hat{\Psi}$ (For GT, we show the curve of white-light LED, which is used to obtain GT RGB images).}
\label{fig:impact}
\vspace{-2mm}
\end{figure*}

\begin{table}[t]
\centering
\caption{Quantitative results on different values for visibility threshold $\hat{\Psi}$. The best results are in \bestscore{red} whereas the second best are in \secondscore{blue}.}
\begin{tabular}{lcccc}
\toprule
Datasets            & $\hat{\Psi}$ & SSIM $\uparrow$   & PSNR $\uparrow$  & LPIPS $\downarrow$  \\
\midrule
\multirow{3}{*}{Ours}      & 10    & 0.8383 & 24.12 & 0.1129 \\
                    & 250          & \secondscore{0.8779}      & \secondscore{25.64}     & \secondscore{0.0919}      \\
                    & 500          & \bestscore{0.9326}      & \bestscore{29.07}     & \bestscore{0.0351}      \\
\midrule
\multirow{3}{*}{TokyoTech} & 10    & 0.7938 & 22.08 & 0.1378 \\
                    & 250          & \secondscore{0.8495}      & \secondscore{23.77}     & \secondscore{0.0920}  \\
                    & 500          & \bestscore{0.9375}      & \bestscore{31.85}     & \bestscore{0.0355}  \\
\bottomrule
\end{tabular}
\label{tab:impact}
\vspace{-3mm}
\end{table}

As has been introduced in Sec.~\ref{sec:MVM}, our model accepts a visibility upper-bound threshold $\hat{\Psi}$ during the spectrum optimization process. This threshold can be changed according to the desired level of visibility and largely affects the final results, without destroying the validity of our method. In this section, we further discuss the impact of this visibility threshold on our model. Specifically, we set different values for the visibility threshold $\hat{\Psi}$, designing the optimal spectrum under each value, and compare their restoration results. The visual and numerical results are reported in Fig.~\ref{fig:impact} and Tab.~\ref{tab:impact}. We can see that as the value of $\hat{\Psi}$ grows, the restoration results of our model become better since more VIS information is covered. The results also imply that we can trade visibility friendliness for restoration performance by setting different value for $\hat{\Psi}$, making our model applicable to a wider range of application scenarios.

\noindent \textbf{Shape of the optimal curves.} As shown in Fig.~\ref{fig:realize}, the scotopic visibility function roughly covers \SI{400}{nm}-\SI{600}{nm}, with a peak value of 1700. High visibility wavelengths have a higher `cost' per intensity than wavelengths with lower visibility. As a result, the model chooses to approach from `sides' with low visibility to the `center' (\SI{500}{nm}) of the visibility curve during the design process. From Fig.~\ref{fig:impact}, we can see that the model still tends not to use around \SI{500}{nm} even when $\hat{\Psi}$=500 since the `cost' is too high. An intuitive thought is that the designed curve should distribute low intensities that fulfill the visibility constraint to a wide VIS range to provide information of different wavelengths. This is, however, not always feasible because of the \textit{noise interference}, especially under strict visibility constraints (\eg, $\hat{\Psi}$=10). Specifically, when the intensity of light becomes very low to cover the high visibility range, the structures and colors may suffer from severe degradation due to \textit{low signal-to-noise ratio}. This makes the model prefer wavelengths that can achieve relatively high intensity under strict visibility constraints, instead of distributing low intensities to a wide VIS range.

\section{Conclusion}
In this paper, we proposed a visibility-constrained wide-band illumination spectrum design (VCSD) model for Seeing-in-the-Dark. Our key insight is to incorporate the quantified visibility constraint implied by the human vision system into the optimization process. By modeling the image formation process in the VIS-NIR range, the optimal multiplexing of a wide range of LEDs is designed in a fully automatic manner, while fulfilling the visibility constraint. We also collected a substantially expanded VIS-NIR hyperspectral image dataset for experiments by using a customized 50-band filter wheel. Experimental results show that the task can be significantly improved by using the optimized wide-band illumination than using NIR only. Further analysis also proved the generality and flexibility of our model to deal with different visibility thresholds.

Although narrow band LEDs are cost effective, they might not be the most appropriate choice when the purpose is to recover high-fidelity visible color, due to the scale-down operation implied by the visibility constraint. Our future work is to allow more flexible illumination design by using wide band fluorescent dyes or customizing thin-film interference filters. 

\section*{Acknowledgement}
This research was supported in part by JSPS KAKENHI Grant Numbers 22H00529, 20H05951, and ROIS NII Open Collaborative Research 2023-23S1201.

{\small
\bibliographystyle{ieee_fullname}
\bibliography{egbib}
}

\end{document}